# AI-enabled exploration of Instagram profiles predicts soft skills and personality traits to empower hiring decisions

**Mercedeh Harirchian[a], Fereshteh Amin[a,*], Saeed Rouhani[b], Aref Aligholipour[c], Vahid Amiri Lord[d]**

[a] *Department of Business Management, University of Tehran, Tehran, Iran*

[b] *Department of Information Technology, University of Tehran, Tehran, Iran*

[c] *Department of Financial Management, University of Tehran, Tehran, Iran*

[d] *Department of Computer Engineering, Damavand Branch, Islamic Azad University, Damavand, Iran*

**Declaration of interest**

None.

**Credit author statement**

**Mercedeh Harirchian:** conceptualization, methodology, Software, Formal analysis, Investigation, Resources, Data Curation, Writing; **Fereshteh Amin:** conceptualization, methodology, Validation, Supervision; **Saeed Rouhani:** conceptualization, methodology, Validation; **Aref Aligholipour and Vahid Amiri Lord:** Investigation, Data Curation

[*] Corresponding author. Department of Business Management, University of Tehran, 16th Azar St., Enghelab Sq., Tehran, Iran, Asia

*E-mail addresses:* m.harirchian@ut.ac.ir (M. Harirchian), famin@ut.ac.ir (F. Amin), srouhani@ut.ac.ir (S. Rouhani), aligholipour@ut.ac.ir (A. Aligholipour).

**Abstract**

It does not matter whether it is a job interview with Tech Giants, Wall Street firms, or a small startup; all candidates want to demonstrate their best selves or even present themselves better than they really are. Meanwhile, recruiters want to know the candidates' authentic selves and detect soft skills that prove an expert candidate would be a great fit in any company. Recruiters worldwide usually struggle to find employees with the highest level of these skills. Digital footprints can assist recruiters in this process by providing candidates' unique set of online activities, while social media delivers one of the largest digital footprints to track people. In this study, for the first time, we show that a wide range of behavioral competencies consisting of 16 in-demand soft skills can be automatically predicted from Instagram profiles based on the following lists and other quantitative features using machine learning algorithms. We also provide predictions on Big Five personality traits. Models were built based on a sample of 400 Iranian volunteer users who answered an online questionnaire and provided their Instagram usernames which allowed us to crawl the public profiles. We applied several machine learning algorithms to the uniformed data. Deep learning models mostly outperformed by demonstrating 70% and 69% average Accuracy in two-level and three-level classifications respectively. Creating a large pool of people with the highest level of soft skills, and making more accurate evaluations of job candidates is possible with the application of AI on social media user-generated data.

**Keywords:**

Hiring Decision, Soft Skill, Competency, Personality, Social media, Artificial Intelligence

**1. Introduction**

Before the digital era and the appearance of social media, humans had never disclosed their thoughts, desires, preferences, and decisions, to this extent, entirely free for others. With the increased general use of social media platforms, academics and businesses are exploring ways to utilize social media in different forms of human subjects(Gelinas et al., 2017; Van Iddekinge, Lanivich, Roth, & Junco, 2016). Multiple studies infer that many organizations are using the Internet and social media, in particular, to search for information about job applicants(Caers, 2011; Van Iddekinge et al., 2016; Willey, White, Domagalski, & Ford, 2012). For example, a survey of 1404 recruiting and human resources professionals revealed that in the recruiting process, only 4% of them do not use social media. 92% of recruiters do use social media, and 4% are not sure(Jobvite, 2015). Careerarc (2021) by surveying 1156 respondents showed that social and professional networks are the most used resource to recruit talent in 2021 (92%). A more recent survey by Jobvite (2021) of 817 recruiters in the U.S. showed that Facebook is the most-used channel for recruitment efforts (with 68% taking advantage) followed by LinkedIn (65%). However, LinkedIn is the best social channel for sourcing the highest quality candidates with 53% of recruiters using it(Jobvite, 2021). Instagram's usage for recruiting has experienced a considerable increase of 28% since 2017, driven chiefly by recruiters under 50. In 2021, 46% of recruiters investigated in recruiting efforts on Instagram(Jobvite, 2021). Another annual survey by SocialTalent of more than 2000 recruiters showed that in 2017, social media was their number one option for finding all-star candidates (34%). LinkedIn was the most valuable resource to a recruiter. Facebook and Instagram were getting more attention, while Twitter was going down regarding recruiting efforts(SocialTalent, 2017). Social sleuthing is standard in recruiting(Jobvite, 2018). Based on Jobvite 2021 survey 30% of recruiters leverage social media to learn more about job candidates(Jobvite, 2021). When recruiters do go digging on social media, they view some factors most positive: engagement in local/national organization groups (60%), examples of written or design work (58%) and mutual connections (36%)(Jobvite, 2018). However, spelling and grammar mistakes (45%), references to marijuana (40%), alcohol consumption (39%), political posts (30%), and pictures of body showing skin (24%) give recruiters pause when researching candidates (Jobvite, 2021). A Belgian study states that a remarkable 40.2% of respondents believe the profile picture provides a signal on the applicant's level of extraversion, and 43.4% see a correlation with maturity(Caers, 2011). A career coach explains "a resume can tell recruiters candidates' qualifications, but a social media profile can help them determine candidates' personality type and if the candidate would be a good fit for company culture(Kramer, 2018)." Another survey by CareerBuilder among 1012 human resource managers in 2018 represents several reasons for hiring a candidate based on the content they found on social media. For example, they saw Job candidate's background information supported their professional qualifications for the job (37%), Job candidate was creative (34%), Job candidate's site conveyed a professional image (33%), Job candidate showed a wide range of interests (31%), could see a good fit within the company culture (31%) and Job candidate had excellent communications skills (28%) (CareerBuilder, 2018). Despite all this, LinkedIn's report on global recruiting trends in 2018 says that traditional techniques in hiring are still popular and effective(Spar, Pletenyuk, Reilly, & Ignatova, 2018). Self-report personality questionnaires, structured and behavioral interviews are widely used, but the survey shows traditional interviews fall short especially in sizing up candidates' soft skills and weaknesses(Spar et al., 2018). Levashina & Campion (2007) show that over 90% of undergraduate job candidates fake during employment interviews. It is hard to evaluate grit in a candidate or spot disorganization by merely having a chat(Melchers, Roulin, & Buehl, 2020; Spar et al., 2018). Therefore, applying indirect methods in assessing candidates seems to be necessary, as many recruiting professionals have already recognized this need and made efforts.

Impromptu evaluations of people on social media contain significant information relevant to their character and competence, but there is also a high level of error, mainly when only a single person makes such judgments (Connelly & Ones, 2010; Van Iddekinge et al., 2016). An efficient way to cut bias out of the social media screening process and to support decisions based on objective data analysis is to use artificial intelligence (AI). Although many studies have focused on the algorithmic fairness of machine learning itself, in response to observed bias in predictions (Choi, Farnadi, Babaki, & Van den Broeck, 2020; Mohammadi, Sivaraman, & Farnadi, 2022), some studies proved that computer-based judgments are more accurate than human judgments and can be a surrogate for human raters in employee selection (Campion, Campion, Campion, & Reider, 2016; Youyou, Kosinski, & Stillwell, 2015). For instance, a study showed that computer personality predictions based on Facebook Likes are more reliable than those made by the participants' Facebook friends and with enough likes, even more than their spouses(Youyou et al., 2015). Moreover, AI-based approaches in recruitment and selection can be executed explicitly based on resumes and interviews or implicitly formed on digital footprints, for example. Most research so far has focused on explicit ways of screening and assessing candidates like resume screening(Qin et al., 2020), video interview assessment (Asan & Soyer, 2022; Batrinca, Mana, Lepri, Pianesi, & Sebe, 2011; L. Chen et al., 2017; Rasipuram & Jayagopi, 2016), written content scoring (Campion et al., 2016), and some other recent works have investigated candidate perceptions and intentions regarding this procedure (Acikgoz, Davison, Compagnone, & Laske, 2020; Mirowska, 2020; Noble, Foster, & Craig, 2021; Wesche & Sonderegger, 2021). Resumes and interviews can be subject to faking and impression management. The present study addresses this gap by using digital footprints as an implicit data source to assess job candidates while not under pressure to manage their impressions or fake their behavior in front of recruiters and hiring managers.

Some studies have examined legal concerns and ethical issues as well as the advantages of using social media profiles for screening job candidates (Hunkenschroer & Kriebitz, 2022; Jeske & Shultz, 2016; Willey et al., 2012). However, little research has examined the validity of using social media for this practice, and the existing ones usually lack sound theoretical foundations (El Ouirdi, El Ouirdi, Segers, & Pais, 2016; Van Iddekinge et al., 2016). One exception is a study that proves values and personality traits predicted based on social media data can help fit individuals into their ideal job(Kern, McCarthy, Chakrabarty, & Rizoiu, 2019). Nevertheless, this study uses the IBM Watson Personality Insights system, a commercial service, to infer social media users' personality characteristics and values which means this study does not directly examine the validity of the practice. Also, the perspective of the study is different because it aims to help people pursue the most matching jobs. From a recruiter's perspective, the position and its prerequisites are more critical, and the candidates are scored, not the job positions.

In the psychology area, there is a bunch of research that apply machine learning algorithms to examine the predictability of personality traits through social media profiles(Farnadi et al., 2016; Kosinski, Stillwell, & Graepel, 2013; Quercia, Kosinski, Stillwell, & Crowcroft, 2011). In addition to personality, some other influential features in hiring decisions are intelligence(Kosinski et al., 2013),values(Kern et al., 2019), interpersonal competency(Jenkins-Guarnieri, Wright, & Hudiburgh, 2012), use of addictive substances(Kosinski et al., 2013), clinical symptoms of psychiatric disorders(Rosen, Whaling, Rab, Carrier, & Cheever, 2013), personal values(J. Chen, Hsieh, Mahmud, & Nichols, 2014), and opinion leadership(Xu, Sang, Blasiola, & Park, 2014) which research shows they are validly predictable with the data from social networking sites. As Lievens and Van Iddekinge(2016) suggest, the next step is to use new approaches to increase the probability of providing accurate job-related information on candidates' knowledge, skills, abilities, and other characteristics (KSAOs).

However, to the best of our knowledge, no work predicts a wide range of job-related behavioral competencies (soft skills) to predict person-job fit through social profiles, as an unobtrusive method, with the help of computational models; this study addresses this gap by predicting 16 soft skills like teamwork, negotiation, persuasiveness and quality orientation automatically through Instagram profiles. We also predict personality traits based on the Big Five model.

Surveys confirm that soft skills are in great demand by employers worldwide, but businesses often stumble in finding employees with high levels of these skills(Cunningham & Villaseñor, 2016). A study on the importance and value of soft skills found that soft skills training, like teamwork and problem-solving, increases productivity and retention by 12 percent and provides a 250 percent return on investment(Adhvaryu, Kala, & Nyshadham, 2019).

So far, most related research projects have concentrated on Facebook(Ortigosa, Carro, & Quiroga, 2014; Youyou et al., 2015), Twitter(Golbeck, Robles, Edmondson, & Turner, 2011; Quercia et al., 2011), YouTube(Biel, Teijeiro-Mosquera, & Gatica-Perez, 2012) or LinkedIn(Faliagka, Tsakalidis, & Tzimas, 2012) but this study select Instagram as one of the fastest growing social network sites globally. Using an Iranian sample is another contribution of this investigation to indicate that despite various cultural implications from western nations, Iranian users with similar soft skills and personality traits also show similar behavioral patterns on social media.

In this work, four machine learning algorithms (Deep Learning, Generalized Linear Model, Logistic Regression, and Random Forest), as well as a baseline algorithm (Decision Tree), are conducted to predict 16 in-demand soft skills and 5 personality traits based on the Big Five personality model in 2-level and 3-level classifications for Iranian Instagram users. Following popular accounts, profile and post-related parameters, and demographic characteristics are three groups of features which models were built upon for each of the 21 target traits. Correlation analysis is also provided for a better interpretation of the predictors.

## 2. Background and Related Work

*2.1. person-job fit and competencies*

Person-job fit centers on narrowing the applicant pool to those who are best qualified for the position opening in question(Norton, 2008). This qualification is mostly described as the knowledge, skills, abilities, and other factors (KSAOs) or competencies required by the job(Bohlander & Snell, 2010; Catano, 2009). Boyatzis popularized the term "competency" and defined it as a combination of a motive, trait, skill, aspect of one's self-image or social role, or a body of relevant knowledge(Catano, 2009). This definition left much room for debate; therefore, a consensus among HR experts has not been reached yet(Edwards, Scott, & Raju, 2003). Nevertheless, different definitions of competency all contain three elements. First, most suggest that competencies are the KSAOs that underlie effective and successful job performance; second, the KSAOs must be observable or measurable; and third, the KSAOs must distinguish between superior and other performers(Catano, 2009). Accordingly, competencies are measurable attributes that distinguish outstanding performers from others in a defined job context.

There are different types and levels of competencies that are categorized in several ways. In a common category, competencies are divided into three types: behavioral competencies, technical competencies, and National and Scottish Vocational Qualifications (NVQs/SNVQs)(Armstrong, 2006). This study is based on behavioral competencies which define behavioral expectations, i.e., the type of behavior required to deliver results under such headings as communication, initiative, and teamwork. They are sometimes known as "soft skills"(Armstrong, 2006). In this work, 16 behavioral competencies were selected from competency dictionaries: innovation, negotiation, communication, gaining commitment, sales ability, strategic decision making, quality orientation, stress tolerance, initiative, work standards, teamwork, decision making, energy, planning and organizing, follow-up, and continuous learning. They are characterized in **Table** *1* (*Harvard University Competency Dictionary*, 2008):

**Table 1**
Overview of the 16 behavioral competencies (soft skills) based on Harvard University Competency Dictionary

| Trait | Description |
| --- | --- |
| Innovation | Generating innovative solutions in work situations |
| Negotiation | Effectively exploring alternatives and positions to reach outcomes that gain the support and acceptance of all parties |
| Communication | Clearly conveying information and ideas to individuals or groups in a manner that engages the audience |
| Gaining Commitment | Using appropriate interpersonal styles and techniques to gain acceptance of ideas or plans |
| Sales Ability/Persuasiveness | Using appropriate communication methods to gain acceptance of a product, service, or idea from prospects and clients |
| Strategic Decision Making | Obtaining information, identifying key issues, and committing to a course of action to accomplish a long-range goal |
| Quality Orientation | Accomplishing tasks by considering all areas involved, no matter how small |
| Stress Tolerance | Maintaining stable performance under pressure or opposition |
| Initiative | Taking prompt action to accomplish objectives; being proactive |
| Work Standards | Setting high standards of performance for self and others |
| Teamwork | Developing and using collaborative relationships to facilitate the accomplishment of work goals |
| Decision Making | Identifying problems and opportunities; comparing data from different sources; developing appropriate solutions; taking proper action that is consistent with available facts |
| Energy | Consistently maintaining high levels of activity or productivity; sustaining long working hours when necessary |
| Planning and Organizing | Establishing courses of action for self and others to ensure that work is completed efficiently |
| Follow-Up | Monitoring the results of delegations, assignments, or projects, considering the capabilities of the assigned individual and the characteristics of the assignment or project. |
| Continuous Learning | Actively identifying and taking advantage of learning opportunities; using newly gained knowledge and skill on the job and learning through their application |

*2.2. The Big Five Personality Inventory*

The system that appears to have won the vote of most personality psychologists is the Five-Factor Model, also introduced as the Big Five personality traits(Chamorro-Premuzic, 2016). Despite much recent consensus about a possible five-factor model for personality, that sometimes leads researchers to use the term, "The Big Five," it would be more fitting to say the big fives, since there is no single set of identical dimensions agreed upon by all researchers(Matthews, Deary, & Whiteman, 2003). In this work, we adopt the five-dimensional model of Costa and McCrae(Costa & McCrae, 1985) based on the vast amount of empirical research that has been done by them and others.

Matthews and et al.,(2003) point out that the development of the five-factor model of Costa and McCrae has been directed somewhat by rational and somewhat by statistical concerns.

Five dominant personality traits of the Five-Factor taxonomy are, namely Neuroticism, Extraversion, Openness to Experience, Agreeableness, and Conscientiousness. Neuroticism can be defined as the inclination to experience negative emotions, notably anxiety, depression, and anger. Extraversion indicates the high activity, the experience of positive emotions, impulsiveness, assertiveness, and a tendency toward social behavior. Openness to Experience represents the inclination to engage in intellectual activities and experience new sensations and ideas. This factor is also mentioned as creativity, intellect, and culture. Agreeableness refers to friendly, considerate, and modest behavior. Finally, Conscientiousness is associated with proactivity, responsibility, and self-discipline(Chamorro-Premuzic, 2016). Many studies have shown the relationship between Big Five personality factors and job performance. Conscientiousness (as a positive predictor) and neuroticism (as a negative predictor) in most cases have a strong relationship with overall job performance(Barrick & Mount, 1991; Hurtz & Donovan, 2000; Salgado, 1997). For other personality factors, the correlation with performance varies by occupation(Barrick & Mount, 1991). When Employers are assessing a candidate's future performance, they consider people-skills and motivation as much as their ability to perform a job. This perspective makes personality a key feature of employability(Hogan, Chamorro-Premuzic, & Kaiser, 2013).

*2.3. AI and recruitment and selection*

So far, most of the AI-based approaches in recruitment and selection have been executed based on automatic resume screening and video interview assessment which are all explicit forms of assessing job candidates. For instance, Qin et al. (2020) aimed to deal with the problem of Person-Job Fit. They focused on measuring the matching degree between job requirements in a job posting and the candidates' experiences in a resume. They suggested a novel end-to-end Topic-based Ability-aware Person-Job Fit Neural Network (TAPJFNN) framework to predict Person-Job Fit based on historical recruitment records. They described how to use their framework for talent sourcing and job recommendation.

There are more examples focused on video interview assessment like an early work by Batrinca, Mana, Lepri, Pianesi, & Sebe (2011). This study addresses the automatic detection of the Big Five personality traits from short self-presentation videos. Researchers analyzed the significance of 29 acoustic and visual, non-verbal features. Based on this work Conscientiousness and Emotional Stability/Neuroticism are the best identifiable characteristics. Another work is by Asan & Soyer (2022) who presented a method that can transform, weight, integrate, and rank automated Asynchronous Video Interviewing (AVI) assessments developed through machine learning algorithms. The suggested approach provides a completely data-driven personnel selection process. Chen et al. (2017) gathered 1891 monologue job interview videos from 260 online employees. Videos were annotated for personality and hiring recommendation scores. The researchers first used clustering to convert continuous audio/video analysis output to discrete pseudoword documents and then applied text classification techniques to process speech content, prosody, and facial expressions. The results showed that personality traits such as openness, conscientiousness, extraversion, agreeableness, and emotional stability were predictable with an F-measure of 0.8 or better, while predicting hiring recommendation score was harder with an F-measure of 0.6. Finally, Rasipuram & Jayagopi (2016) targeted communication skills in an interface-based interview setting. They collected audio-visual recordings of 106 participants and automatically analyzed audio and visual behavioral cues such as prosodic, speaking activity-based, and facial expression cues. Three independent judges were employed to derive ground truth labels. A proposed framework automatically predicts the communication skills rating using regression and classification models. Prediction accuracy was 56% when labels were not grouped. Accuracy for Below Average Task using SVM with RBF kernel was 80%.

In a different attempt to use AI in hiring practices, Campion, Campion, Campion, & Reider (2016) train a computer program simulating a human rater to score accomplishment records. An accomplishment record is an achievement people attained in the past that demonstrate they have the competency required to perform the job for which they have applied (Campion et al., 2016). This investigation showed that the computer program could simulate a human rater scoring ARs and produce scores as reliable as those of a human rater with construct validity. Additionally, the computer scores showed no race or gender discrimination. The study indicated that computer scoring could result in considerable cost savings in scoring ARs. Finally, the researchers concluded that replacing any of the three employed human raters with the computer program made no difference due to the small number of different hiring decisions.

*2.4. Job-Related Features and Social Media*

Studies on predicting personality, behavioral features and general mental ability in social media which are all essential features in staffing decisions can be divided into two categories: The ones conducted with the aim of employee selection, and the ones directed by researchers in other areas such as psychology and computer science and not directly with a recruitment purpose. Studies on hiring decisions either evaluated the ability of HR professionals to predict job performance or withdrawal(Kluemper, Rosen, & Mossholder, 2012; Van Iddekinge et al., 2016) or assessed the pros and cons of using such data in the staffing process and the legal and ethical concerns(Gelinas et al., 2017; Jeske & Shultz, 2016).

However, some researchers operated AI to evaluate the ability of such tools to predict psychological elements and only a few of them to appraise the employability of job seekers based on their SM profiles(Faliagka et al., 2012; Kern et al., 2019). The fact that there have been few studies on using AI in processing SM data to help make personnel selections, indicates that it is still not established among HR researchers in spite of its widespread use in the industry.

Among those studies which employed human professionals to estimate job performance and personality, we can mention Kluemper et al.'s work(2012). They used Facebook profiles to rate students' personality and compatibility for hiring. A few research assistants reviewed the profiles. For a subset of the students, the ratings were correlated with supervisor ratings of job performance. Results showed that ratings of emotional stability and agreeableness based on Facebook profiles correlated significantly with performance, while ratings of the other personality factors did not. Another research(Van Iddekinge et al., 2016) with a similar process examines the validity of these profile ratings by recruiters with two critical criteria: criterion-related validity and subgroup differences. Recruiters from various organizations reviewed the Facebook profiles of college students who were applying for full-time jobs. Recruiters provided evaluations, and then researchers followed up with applicants in their new jobs. Results showed that recruiter ratings of applicants' Facebook information were not related to supervisor ratings of job performance, turnover intentions, and actual turnover. In addition, there was evidence of bias in favor of female and white applicants in Facebook ratings.

The study of Faliagka et al. (Faliagka et al., 2012) is among those rare ones that aim to use AI to know whether it can make hiring decisions a smoother process with the help of SM data. This research leverages machine learning algorithms to solve the candidate ranking problem. The job applicants are being evaluated in an online recruitment system. The system implements automated candidate ranking based on individual criteria, i.e., education, work experience, loyalty, and extraversion. Most of the criteria are derived from the applicant's LinkedIn profile, and extraversion is extracted from applicant's social presence in personal blogs using linguistic analysis. Analytical hierarchy process (AHP) is used to rank applicants. The researchers conducted the e-recruitment system in a real-world recruitment scenario, and it was found that the system performed consistently compared to human recruiters.

With a slightly different perspective Kern, McCarthy, Chakrabarty, & Rizoiu (2019) provide evidence that based on linguistic information unobtrusively accumulated through social media, the most fitting professions for individuals can be predicted. The researchers illustrate distinguishing psychological profiles for professions Based on 128,279 Twitter users representing 3,513 occupations. Parallel disciplines cluster together, indicating specific groups of jobs. Formed on values and personality profiles people are predicted to be well suited for one group of jobs.

As previously mentioned, there is another group of papers which have some contributions for research in domain of human behavior prediction by computer, but they do not directly focus on employee selection. In this group, we can refer to Chen and Hsieh's study (2014). They analyzed people's values and their word use on Reddit. A sample of 799 Reddit users contributed to this survey by answering surveys and providing their text on Reddit to researchers. Chen and Hsieh found categories of words that are indicative of user's value orientation. They further provided an explorative report on human values prediction based on text analysis. In another work, Xu et al. (2014) explore the predictors of opinion leadership on Twitter. They used user-generated content to measure user characteristics. The results show the feasibility of this process and demonstrate that higher connectivity and issue involvement are the critical characteristics of those Twitter users who are highly influential.

Some studies focus on psychological elements to examine the practicality of extracting them from technology use behavior. Among these studies, we can mention Rosen et al.'s work (2013) which tries to figure out whether the use of specific technologies or media, technology-related anxieties, and technology-related attitudes would predict clinical symptoms of six personality disorders and three mood disorders. Results showed that each disorder had a unique set of predictors. Facebook general use, friendship, and impression management were identified as significant predictors. For instance, more Facebook friends predicted more clinical symptoms of bipolar-mania, narcissism and histrionic personality disorder. In this study, all sorts of technology use behavior, like the frequency of Facebook use, were directly asked from participants through questioners and researchers did not have any access to the profiles of the participants. Another study suggests that interpersonal competency can be predicted by the intensity of Facebook use since greater intensity of using Facebook was associated with perceptions of decreased interpersonal competency at initiating relationships(Jenkins-Guarnieri et al., 2012).

*2.5. Personality and Social Media*

A plethora of studies is dedicated to personality prediction through social media. Quercia et al. (2011) and Golbeck et al. (2011) allegedly did some of the first studies in this realm. Quercia et al. (2011) found that both popular users and influentials on Twitter, are extroverts and emotionally stable. Popular users are more open to new experiences, while influentials are more inclined towards being conscientious. Based upon three counts publicly available on profiles, they showed a way of predicting a user's personality traits with a root-mean-squared error below 0.88 on a [1,5] scale. Golbeck et al. (2011) did a similar work on Twitter and with the application of machine learning techniques predicted personality of the users based on five-factor model. They predicted scores on each of the five personality traits to within 11% - 18% of their actual value. They did another research on Facebook in which they trained two machine learning algorithms and predicted each of the five personality traits to within 11% of its actual value (Golbeck, Robles, & Turner, 2011).

In a more recent work, with a much greater sample (over 58,000 volunteers), Kosinski et al. (Kosinski et al., 2013) show that highly sensitive personal attributes like sexual orientation, religious and political views, personality traits, intelligence, use of addictive substances, parental separation, etc. can be easily predicted with the help of Facebook Likes. Among personality traits of Big Five model, openness prediction accuracy was more acceptable as it was close to the test-retest accuracy of the test. In another work scholars compared the accuracy of human and computer-based personality judgment (Youyou et al., 2015). The results of analyzing a sample of 86,220 volunteers' Facebook likes and their answers to a 100-item personality questionnaire showed that computer predictions are more accurate. Only a volunteer's spouse can judge them slightly more accurately than a computer.

Ortigosa et al. (2014) developed a Facebook application to collect information about the personality traits of more than 20,000 users. By analysis of user interactions within Facebook along with the knowledge of their personality traits, researchers were able to infer the personality of users automatically. The results show that the classifiers have a high level of accuracy.

In a comparative study, Farnadi et al. (2016) perform an analysis on a varied set of social media websites: Facebook, Twitter and YouTube. In answering the question that whether all personality traits of a given user should be predicted at once or separately, they found that multivariate regression learners often outperformed the univariate regression ones, but the differences between univariate and multivariate models were not significant. In comparing different features, they found that the Linguistic Inquiry and Word Count tool (LIWC) feature set outperformed others in predicting personality traits for the YouTube and Facebook datasets.

According to the introduced studies, many recent papers have facilitated human resources functions in hiring practices by using artificial intelligence to lessen handiwork, increase speed, and reduce costs. Moreover, many other studies have automatically predicted behavioral characteristics such as personality and intelligence based on digital footprints like social media activities using artificial intelligence and machine learning algorithms.

In this article, by combining these two approaches, we attempt to unobtrusively evaluate job candidates using Instagram profiles based on job-market in-demand soft skills which have been less investigated by previous researchers, and the personality traits, automatically using machine learning algorithms.

## 3. Method

We targeted 21 in-demand soft skills and personality traits, adopted them from the literature, and, based on them, supplied an online questionnaire to collect self-reported data on targeted traits. Participants were asked to provide their Instagram account username. Some volunteers offered us their usernames. Next the content of public Instagram profiles, including profile and post data (caption, comments, hashtags, etc.) and followers and followings lists, were collected employing web crawlers. We then applied several data mining algorithms to the cleaned, uniformed data in order to see whether we could predict individuals' soft skills and personality traits in 2-level and 3-level classifications.

The following list of Instagram users is a special predictor in this study. Popular Instagram accounts belong to musicians, actors, actresses, sports personalities, TV personalities, models, magazines, brands, social media platforms, favorite websites, and bloggers. The following behavior is typically an indicator of interest, association, or taste. Most popular and mutually followed accounts between Iranian research volunteers were used in building the model along with post-related parameters (Average post likes, Hashtag usage ratio, etc.) and demographic characteristics.

### 3.1. Data collection and Procedure

### 3.1.1. Online Questionnaire

We collected the data in two steps. First, we provided online soft skills and personality traits questionnaires through an instant messaging application. Then, we crawled the Instagram profile data of the respondents to the questionnaires. Due to the high popularity of the Telegram application in Iranian society and the demographic diversity of the Telegram users, the research questionnaire was provided to the participants via the Telegram robot. To encourage greater participation, the personality and competency test results were provided at the end of the test to participants who had fully answered the questions. Finally, 842 people responded to the questionnaire in about 50 days. The questionnaire consisted of 4 sections:
- Demographic questions: gender, age, education and occupation
- Behavioral Competency (soft skills) Questionnaire: This questionnaire was designed by the researchers based on the Harvard Competency Dictionary and had 54 independent items.
- NEO-FFI (NEO Five-Factor Inventory) Questionnaire: This questionnaire was designed by Costa and McCrae (2004) and had 60 items.
- The last question about Instagram account usernames (Participants were able to either answer or skip this question)

The Behavioral Competency Questionnaire measures the 16 competencies selected from the full list of 42 competencies of the Harvard Competency Dictionary. To assess the validity of the questionnaire variables, we conducted content validity. Moreover, we used Cronbach's alpha scale to estimate the reliability of the scales. The competencies yielded Cronbach's alphas from 0.6 to 0.88. Based on the scores of the first 20 participants, we calculated the average and standard deviation of each variable to send the result to the participants automatically. The score range of each competency was divided into three sections: high, medium, and low. The high-scoring, mid-scoring, and low-scoring competencies were named strengths, improvable features, and weaknesses, respectively. For each of them, a definition was provided. After increasing the number of participants during the two updating stages, new mean and standard deviation were obtained, and the robot code was modified in the calculation section.

NEO Five-Factor Inventory is reported to possess adequate internal consistencies ranging from 0.7 to 0.86 (McCrae & Costa Jr, 2004). To calculate the results of each individual in the robot and to present it automatically, it was necessary to use the norm of this test and determine the mean scores. In this regard, we used the results of a study with an Iranian sample (Haghshenas, 2014). Characteristics of the participants and descriptive statistics results of our study are given in **Table** *2*.

**Table 2**

(a) Characteristics of 400 participants who disclosed their Instagram usernames. (b) Min, Max, Mean, and SD of soft skills and personality traits for 400 participants.

|  | Missed |  |  |  |  |
| --- | --- | --- | --- | --- | --- |
| (a) |  |  |  |  |  |
| Gender | 0 | Male (143), Female (257) |  |  |  |
| Education | 0 | High school Student (13), Diploma (27), Associate Degree (18), bachelor's Degree (120), Master's Degree (202), Doctorate Degree (20) |  |  |  |
| Occupation | 66 | Employee (243), Self-employed (25), Artist (23), Housewife (14), Student (71), Unemployed (24) |  |  |  |
| Private page | 9 | True (289), False (102) |  |  |  |
|  | Missed | Min | Max | Mean | SD |
| (b) |  |  |  |  |  |
| Age | 25 | 14 | 65 | 28.38 | 6.77 |
| Neuroticism | 0 | 1 | 48 | 21.87 | 9.51 |
| Extraversion | 0 | 7 | 48 | 31.28 | 7.72 |
| Openness to Experience | 0 | 11 | 46 | 29.55 | 5.60 |
| Agreeableness | 0 | 11 | 45 | 30.65 | 5.58 |
| Conscientiousness | 0 | 14 | 48 | 33.94 | 6.88 |
| Innovation | 0 | 2 | 20 | 11.20 | 3.21 |
| Negotiation | 0 | 3 | 20 | 13.56 | 2.67 |
| Communication | 0 | 2 | 20 | 13.48 | 3.23 |
| Gaining Commitment | 0 | 1 | 24 | 16.25 | 4.53 |
| Sales Ability/Persuasiveness | 0 | 0 | 12 | 6.63 | 2.53 |
| Strategic Decision Making | 0 | 1 | 16 | 10.70 | 3.12 |
| Stress Tolerance | 0 | 0 | 12 | 6.95 | 2.79 |
| Initiative | 0 | 2 | 16 | 10.53 | 2.82 |
| Work Standards | 0 | 8 | 32 | 24.93 | 4.38 |
| Decision Making | 0 | 1 | 16 | 10.29 | 3.27 |
| Teamwork | 0 | 0 | 12 | 8.33 | 2.35 |
| Energy | 0 | 0 | 12 | 7.48 | 3.19 |
| Planning and Organizing | 0 | 1 | 12 | 7.65 | 2.81 |
| Follow-Up | 0 | 2 | 16 | 10.05 | 2.64 |
| Continuous Learning | 0 | 6 | 24 | 17.02 | 3.24 |
| Quality Orientation | 0 | 2 | 24 | 15.53 | 4.44 |

*3.1.2. Instagram*

In this study, 842 people responded to the online questionnaire, of which 400 allowed us to know their Instagram account username. Of these, 102 had a private page. Therefore, access to their Instagram data was restricted to three metrics: number of followers, number of followings, and number of posts. We collected the content of public Instagram profiles (289 accounts), including profile and post data (caption, comments, hashtags, etc.) and followers and followings lists using two Instagram crawlers. Instagram-scraper is an open-source crawler developed in Python by Richard Arcega (Arcega, 2017). This tool facilitates downloading the content of public Instagram accounts. Photos and videos, captions, hashtags, locations, number of likes and comments for each post, the content of every comment and the commentators, and post types were collected with this tool. Instabot is another tool which helped us to download all follower and following lists of the users along with their profile pictures and biographies developed by Daniil Okhlopkov (Okhlopkov, 2017). The following list is of great importance in this research because some of the popular accounts in this list show the interests and preferences of the users, just as Facebook likes do.

*3.1.3. Extracted Features*

The data collected from Instagram profiles is shown in **Table 3**. This data can be divided into two categories: first, following popular accounts and, second, profile and post-related parameters such as captions length, hashtags usage ratio, and the number of likes. Following popular accounts is important because, just as Facebook Likes show a positive association, preference, and taste (Youyou et al., 2015), following popular Instagram accounts can also reflect such a relationship. In **Table 3**, "Following each of popular accounts" is not one feature but 830 features that each take the name of a popular Instagram account. We extracted these features from the following lists of participants. The following lists of all participants (400 people) were collected, and then those accounts with more than 50,000 followers were selected. Of these, the accounts followed by at least six research participants were selected as new features. Each of these 830 pages was named as a new feature. Following each of them was displayed by 1 and not following them by 0.

**Table 3**
Min, Max, Mean, and SD of interaction parameters for 400 Instagram users

|  | Missed | Min | Max | Mean | SD |
| --- | --- | --- | --- | --- | --- |
| Number of followers | 9 | 0 | 16687 | 478 | 1491 |
| Number of followings | 9 | 0 | 4357 | 412 | 450 |
| Number of posts | 21 | 0 | 1729 | 109 | 181 |
| Total post likes | 156 | 7 | 281948 | 12741.43 | 27806.03 |
| Average post likes | 156 | 7 | 4300.37 | 96.63 | 282.68 |
| Total post comments | 156 | 0 | 26911 | 1005.91 | 2098.52 |
| Average post comments | 156 | 0 | 64.11 | 8.70 | 7.67 |
| Hashtag usage ratio | 156 | 0 | 1 | 0.32 | 0.30 |
| Number of captions | 156 | 0 | 1130 | 117.27 | 171.40 |
| Average caption length | 156 | 0 | 1084.77 | 168.83 | 158.16 |
| Number of locations | 155 | 0 | 470 | 25.76 | 57.10 |
| Location usage ratio | 155 | 0 | 1 | 0.18 | 0.21 |
| Number of disabled comments | 156 | 0 | 12 |  |  |
| Total caption length | 156 | 0 | 387025 | 23295.25 | 41350.91 |
| Number of image posts | 156 | 0 | 1616 | 123.01 | 194.81 |
| Image posts ratio | 156 | 0 | 1 | 0.88 | 0.20 |
| Number of slide posts | 156 | 0 | 37 | 3.19 | 5.20 |
| Slide posts Ratio | 156 | 0 | 1 | 0.05 | 0.10 |
| Number of video post | 156 | 0 | 143 | 7.66 | 17.55 |
| Video posts ratio | 156 | 0 | 0.68 | 0.05 | 0.09 |
| Average post engagement | 159 | 0.03 | 2.09 | 0.29 | 0.20 |
| Total number of followers of followers | 137 | 0 | 15426258 | 943529 | 1888295 |
| Maximum number of followers of followers | 137 | 0 | 1129745 | 99793.27 | 161853.37 |
| Number of popular followers | 137 | 0 | 35 | 1.81 | 3.99 |
| Following each of popular accounts | 137 | 0 | 1 | | |
| Number of following popular accounts | 136 | 0 | 224 | 23.97 | 36.03 |

## 3.2. Dataset Preparation

We used RapidMiner Studio (Mierswa & Klinkenberg, 2018) for data preparation and predictive analysis. We converted numerical data from questionnaires to categorical data to represent low and high levels in one classification and low, medium, and high levels in another classification, for all the 16 behavioral competencies and five personality traits. For example, we once predicted whether the negotiation competency level for each participant was above or below average. In another attempt, we predicted that the negotiation competency level was either above average plus standard deviation (high), between average plus standard deviation and average minus standard deviation (medium), or below average minus standard deviation (low). We did this because knowing that people are at a high, medium, or low level in each competency and personality trait is more effective in making hiring decisions and examining person-job fit. We need to know that the person we are considering for a marketing position, for example, is entirely creative or moderately creative. One or two points of difference between two people do not have much effect on our decision.

A unique set of features was selected to independently build a predictive model for each of the 21 behavioral competencies and personality traits in 2-level and 3-level classifications. Different feature sets were selected and tested. We conducted correlation analysis to find the most relevant subset of features for each of the 21 target variables with $p < 0.01$.

## 3.3. Classification Models

Initially, we used a variety of algorithms to analyze the data. Due to the favorable primary results of some algorithms, we selected four models: Deep Learning H2O (DL), Generalized Linear Model (GLM), Logistic Regression (LR), and Random Forest (RF). All of these algorithms were used for both 2-level and 3-level classifications, except for logistic regression, which is by nature a binary algorithm.

### 3.3.1. Deep Learning H2O (DL)

Deep learning uncovers complex structures in big data using the backpropagation algorithm to reveal how parameters should change. The Machine uses these internal parameters to calculate each layer's representation from the previous layer's (LeCun, Bengio, & Hinton, 2015). To build the models based on this algorithm, we used 50 hidden layers with rectifier activation. They were reproducible with 1992 local random seed, 10 epochs and samples training per iteration was set to auto-tuning.

### 3.3.2. Generalized Linear Model (GLM)

The GLM unifies various other statistical models, including linear regression, logistic regression, and Poisson regression. This algorithm generalizes linear regression by allowing the linear model to be related to the response variable via a link function and allowing the magnitude of the variance of each measurement to be a function of its predicted value(Vittinghoff, McCulloch, Glidden, & Shiboski, 2011). In this study, we set the family and the solver parameters to AUTO, and the models were reproducible. The maximum number of threads was 1, regularization was used, and the numeric columns were standardized.

*3.3.3. Logistic Regression (LR)*

This algorithm is a simplified arrangement of the Generalized Linear Model, so we set binomial for the family parameter and logit for the link parameter. Using standard linear regression for a binominal output can produce inadequate results. Moreover, the variability of the output should be the same for all values of the predictors for linear regression to be valid. This premise does not match the behavior of a binominal output. So, linear regression is inadequate for binominal data, and Logistic regression fills this gap(LaValley, 2008).

*3.3.4. Random Forest (RF)*

The Random Forest algorithm performs based on the "divide and conquer" principle, which is simple but practical: split data into samples that develop a randomized tree predictor, then combine these predictors together(Biau & Scornet, 2016). The number of trees and maximal depth differed in building various models predicting 21 traits. In this model, the 3-fold cross-validation method is also used in the optimize parameter operator.

*3.4. Validation and Comparison of Algorithms*

The split data approach was chosen with a split ratio of 80/20 so that 80% of the data was used for training and 20% for testing. Accuracy, Area Under Curve (AUC), and precision were calculated to evaluate the performance of 2-level classifications. The validation of 3-level classifications was performed using accuracy and weighted $F_1$ score. Accuracy, precision, AUC, $F_1$ score, and weighted $F_1$ score range from 0 to 1 where higher values signify better models. These metrics are calculated from true positives (TPs), false positives (FPs), false negatives (FNs), and true negatives (TNs). The ratio of correct predictions over all the predictions is called *accuracy*:

$$Accuracy = (TP + TN)/(TP + FP + FN + TN)$$

The fraction of correctly predicted positives to all predicted positives is called *precision*, while the proportion of correctly predicted positives to all actual positives is called recall. They are expressed as:

$$Precision = TP/(TP + FP)$$

$$Recall = TP/(TP + FN)$$

$F_1$ score balances the considerations of precision and recall in one single score, and weighted F1 score is the weighted mean $F_1$ score of the classes. The following formulas can describe them:

$$F\_1 = (2 \times precision \times recall)/(precision + recall)$$

$$Weighted\ F_1 = \frac{\sum_{i=1}^{n} Ni \times F_{1\ class_i}}{\sum_{i=1}^{n} Ni}$$

Where $F_{1\ class_i}$ and $Ni$ are the $F_1$ score of the class $i$ and the number of samples from that class (where $i = 1,…,n$) and $n$ is the sample size.

AUC is the "Area Under the Curve" of the ROC curve (Receiver Operating Characteristic). When 0.5<AUC<1, the classifier is able to detect more numbers of True positives and True negatives than False negatives and False positives.

While accuracy is a very commonly used metric, precision is another critical one that serves the purpose of this study. In the field of human resources and explicitly hiring, selected candidates must have a set of soft skills and personality traits linked with their job role. It highly matters what percentage of the people who are predicted to be in the class of interest (for example, a "high" score in negotiation) truly have this ability. While it is important not to reject people with this trait mistakenly, the cost of accepting people who do not have this trait is higher, so we chose precision over recall.

Furthermore, we acknowledge that accuracy is not a preferable measure of performance for imbalanced data as in the case of 3-level classifications in this study (more data in the Medium class). Hence, we selected weighted F1 score that is a favorable measure of performance for imbalanced multiclass problems.

## 4. Results

### 4.1. Correlation results

The most important correlations between the extracted features and personality and behavioral competency scores are reported in **Table 4**. We used parametric Pearson correlation to compute the correlations and employed the SPSS software environment (IBM Corp., 2019). All the presented correlation results are significant with p < 0.01.

Gender and occupation as demographic features are significantly correlated with some of the personality traits and soft skills. The encoding approach regarding gender usually indicates female with 1 and male with 0. This study used the same system. Occupation and education as multivariate features were also encoded. Seven categories were stablished for occupation and were encoded as Artist = 0, employee = 1, Housewife = 2, Retired = 3, Self-employed = 4, Student = 5, and Unemployed = 6. Six categories of education feature were also encoded: High school student = 0, Diploma = 1, Associate Degree = 2, bachelor's Degree = 3, Master's Degree = 4, Doctorate Degree = 5. Private page is another dichotomous feature that was indicated as 0 when the profile was public and 1 when the profile was private. The significant correlation results show that there is a more significant association between the soft skills and personality traits and following popular accounts. Gender and occupation also showed significant correlations. However, profile and post-related parameters as were introduced in **Table 3** were not among the most significant features. These features were selected in the feature selection process to build some of the prediction models.

Most of the popular Instagram accounts presented in this study are actually popular among Iranian people. However, there are several globally famous Instagram accounts like elonmusk, humansofny, emmawatson, natgeo, ted, adele, nasa, tomhanks, streetartglobe, time, fendi, dolcegabbana, etc. that following them were significantly correlated with the traits. Followers of elonmusk were more likely to show low levels of Neuroticism due to the negative correlation (-0.16), and demonstrate higher levels of stress tolerance (0.16). Following natgeo and paulnicklen (a natgeo contributor) are both positively associated with Openness to Experience (0.18 and 0.21), Innovation (0.17 and 0.19), and Continuous Learning (0.19 and 0.19), which are intuitively comprehensible. Followers of elonmusk were more likely to show low levels of Neuroticism due to the negative correlation (-0.16), and demonstrate higher levels of stress tolerance (0.16). Following natgeo and paulnicklen (a natgeo contributor) are both positively associated with Openness to Experience (0.18 and 0.21), Innovation (0.17 and 0.19), and Continuous Learning (0.19 and 0.19), which are intuitively comprehensible.

**Table 4**

significantly (p < .01) correlated features with personality traits and soft skills

| Trait | | | | | | | | | |
|---|---|---|---|---|---|---|---|---|---|
| Neuroticism | Feature | elonmusk | Gender = F | apnaida | | | | | |
| | Correlation | -0.16 | 0.16 | 0.15 | | | | | |
| Extraversion | Feature | ostad_hosein_alizadeh_fanpage | zibakalamsadegh | mosighi.sonaty | | | | | |
| | Correlation | 0.14 | 0.14 | 0.14 | | | | | |
| Openness to experience | Feature | paulnicklen | humansofny | loffl | mohammadmotamedi | natgeo | | | |
| | Correlation | 0.21 | 0.19 | 0.18 | 0.18 | 0.18 | | | |
| Agreeableness | Feature | Gender = F | so___whatt | mohsenchavoshi | | | | | |
| | Correlation | 0.14 | -0.14 | -0.14 | | | | | |
| Conscientiousness | Feature | 4vafa4 | ahmadmehranfar | kathyshajarian | | | | | |
| | Correlation | -0.18 | -0.16 | -0.16 | | | | | |
| Innovation | Feature | emmawatson | paulnicklen | golfarahani | loffl | natgeo | ted | adele | nasa |
| | Correlation | 0.2 | 0.19 | 0.19 | 0.17 | 0.17 | 0.17 | 0.17 | 0.16 |
| Negotiation | Feature | tahminehmilani | mohsenchavoshi | zynp | tomhanks | | | | |
| | Correlation | -0.18 | -0.18 | -0.16 | 0.16 | | | | |
| Communication | Feature | cafeparagraph | mohammadmotamedi | kayhan_kalhor | | | | | |
| | Correlation | -0.17 | 0.16 | 0.15 | | | | | |
| Gaining Commitment | Feature | Gender = F | saadigram | astad_hosein_alizadeh_fanpage | | | | | |
| | Correlation | -0.17 | 0.16 | 0.15 | | | | | |
| Sales Ability/Persuasiveness | Feature | saadigram | streetartglobe | pronovias | time | | | | |
| | Correlation | 0.17 | 0.16 | -0.16 | 0.15 | | | | |
| Strategic Decision Making | Feature | Gender = F | hodajarrah | ehsaankaramy | | | | | |
| | Correlation | -0.2 | -0.17 | -0.15 | | | | | |
| Stress Tolerance | Feature | Gender = F | astad_hosein_alizadeh_fanpage | mohammadmotamedi | aref_ir | elonmusk | | | |
| | Correlation | -0.21 | 0.18 | 0.18 | 0.17 | 0.16 | | | |
| Initiative | Feature | occupation = employee | 4vafa4 | fendi | | | | | |
| | Correlation | 0.22 | -0.16 | -0.15 | | | | | |
| Work Standards | Feature | occupation = housewife | follower_count | 4vafa4 | | | | | |
| | Correlation | -0.18 | -0.17 | -0.16 | | | | | |
| Decision Making | Feature | Gender = F | astad_hosein_alizadeh_fanpage | dolcegabbana | | | | | |
| | Correlation | -0.17 | 0.16 | -0.15 | | | | | |
| Teamwork | Feature | mahsanemat | topcharter | mcocollection | | | | | |
| | Correlation | -0.15 | -0.14 | -0.14 | | | | | |
| Energy | Feature | Gender = F | gerdeshahrdokan | kathyshajarian | | | | | |
| | Correlation | -0.27 | -0.17 | -0.17 | | | | | |
| Planning and Organizing | Feature | sahardolatshahi | englishpersian | mahnaz_afshar | | | | | |
| | Correlation | 0.15 | 0.14 | 0.14 | | | | | |
| Follow-Up | Feature | sondos_afsaneh.rahimian | fendi | babaktafreshi | | | | | |
| | Correlation | -0.17 | -0.16 | -0.15 | | | | | |
| Continuous Learning | Feature | natgeo | paulnicklen | occupation = housewife | zibakalamsadegh | | | | |
| | Correlation | 0.19 | 0.19 | -0.17 | 0.17 | | | | |
| Quality Orientation | Feature | piugan | mohsenchavoshi | soroushsehat | | | | | |
| | Correlation | 0.17 | -0.17 | -0.16 | | | | | |

Emmawatson, ted, adele, and nasa are among other accounts that have positive correlation with innovation (0.2, 0.17, 0.17, 0.16). Streetartglobe and time were positively related to Sales Ability/Persuasiveness (0.16 and 0.15), whereas pronovias (a wedding dress brand) showed a negative relation with this trait (-0.16). Fendi (a clothing brand) was negatively correlated with both Initiative (-0.15) and Follow-up (-0.16) traits. However, piugan, another clothing brand, showed a positive relation with Quality Orientation. Tomhanks was associated with negotiation in a positive way (0.16, perhaps influenced by his role in the movie Bridge of Spies). Following Golshifteh Farahani (golfarahani) was positively related to innovation trait (0.19). She is one of the few Iranian actresses who has acted in Hollywood movies.

Among popular Iranian accounts, zibakalamsadegh, for instance, which is owned by Sadegh Zibakalam, was an indicator of Extraversion (0.14) and Continues Learning (0.17). Zibakalam is an author and professor of political sciences. Following so___whatt showed negative relation with Agreeableness (-0.14). Ahmadmehranfar (actor) had a negative association with Conscientiousness (-0.16). Tahminehmilani (movie director) and mohsenchavoshi (singer) both displayed a similar negative correlation with the Negotiation trait (-0.18). Following accounts of two singers (mohammadmotamedi and kayhan_kalhor) were positively correlated with the communication trait (0.16 and 0.15), while following a cafe (cafeparagraph) was negatively correlated with this trait (-0.17). Saadigram, an account dedicated to Saadi, a major Persian poet and prose writer of the medieval period, was positively related to Sales ability/Persuasiveness (0.17). He is nicknamed "The Master of Speech" or "The Wordsmith". Following aref_ir (politician), was related to Stress Tolerance in a positive way (0.17). Two Instagram accounts most relevant to solo traveling, mahsanemat (traveler) and topcharter (charter flight service provider), showed negative relations to the Teamwork skill (-0.15 and -0.14). There was a positive relation between following Englishpersian (educational) and the Planning and Organizing trait (0.14). Two of the negative indicators of Quality Orientation were following soroushsehat (director and actor) and mohsenchavoshi (-0.16 and -0.17).

Gender and Occupation among the other demographic features showed more significant relation with soft skills and personality traits. Comparing mean scores for men and women showed that being a female is positively related to Neuroticism (0.16) and Agreeableness (0.14), while negatively correlated with Gaining Commitment (-0.17), Strategic Decision Making (-0.2), Stress Tolerance (-0.21), Decision Making (-0.17), and Energy (-0.27). Among the seven occupation categories, housewives showed more significantly negative relation with Work Standards (-0.18) and Continuous Learning (-0.17). Moreover, for employees mean Initiative score was higher among other occupation categories and showed a positive relation with this soft skill (0.22).

*4.2. Regression Models*

As previously described, four algorithms were used to examine how accurately we can predict 21 self-reported behavioral competencies (soft skills) and personality traits based on Instagram account data. According to the purpose of this study, explained in section 3.2, numerical data obtained from questionnaires was transformed into categorical data to predict high, medium, and low levels of traits in 3-class problems and high and low levels in 2-class problems for each 21 target variables based on unique sets of features. The feature selection process as an important step in increasing the accuracy of the model is described in the next section, followed by the results of applying the models.

*4.2.1. Feature Selection and Building the Models*

There are a number of objectives for feature selection: preparing clean and comprehensible data by identifying relevant features and removing irrelevant ones before training the model, enhancing data-mining performance as it was proven by previous studies(Farnadi, Zoghbi, Moens, & De Cock, 2013; Hall, 1999), and building straightforward and more understandable models(Li et al., 2017). Many feature selection methods have been suggested to obtain the feature subsets in the literature as this procedure has become an indispensable task to apply machine learning algorithms(Kumar & Minz, 2014). We preferred correlation analysis since this practice allows us to understand the underlying impact of each of the features in predicting soft skills and personality traits. Moreover, it does not depend on any specific algorithm as it is filter-based and therefore can produce general results(Farnadi et al., 2016).

For each trait, we chose features with correlation coefficients equal to or more than 0.01 with the class of highest interest (the class with the most frequency either in 2-level classification or 3-level classification) with $p < 0.05$. Then, run several algorithms (DL, GLM, LR, RF) in RapidMiner studio to train and test the models based on the feature set. In the following iterations, we removed some features with the lowest correlation. Then added some of them and pulled some of the others until we reached the optimal accuracy, which also outperforms the baseline model (DT), with at least one of the performing algorithms for each trait. **Fig *1*** and **Fig *2*** illustrate the percentage of feature set categories selected to build the model in the two-level and three-level classifications.

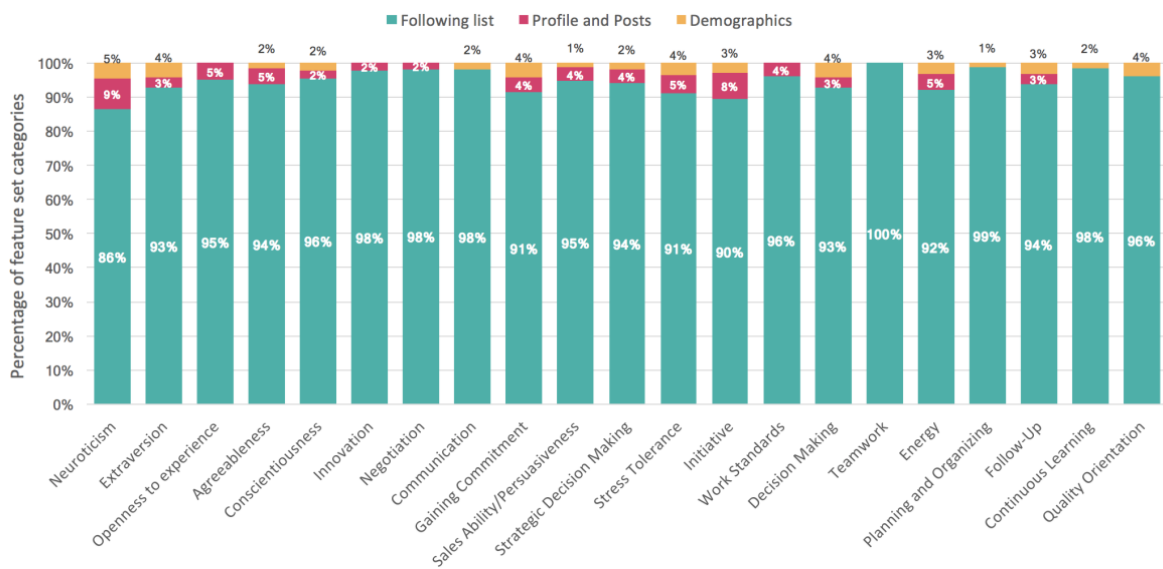

**Fig 1** Percentage of the model input feature set categories after feature selection in the two-level classification

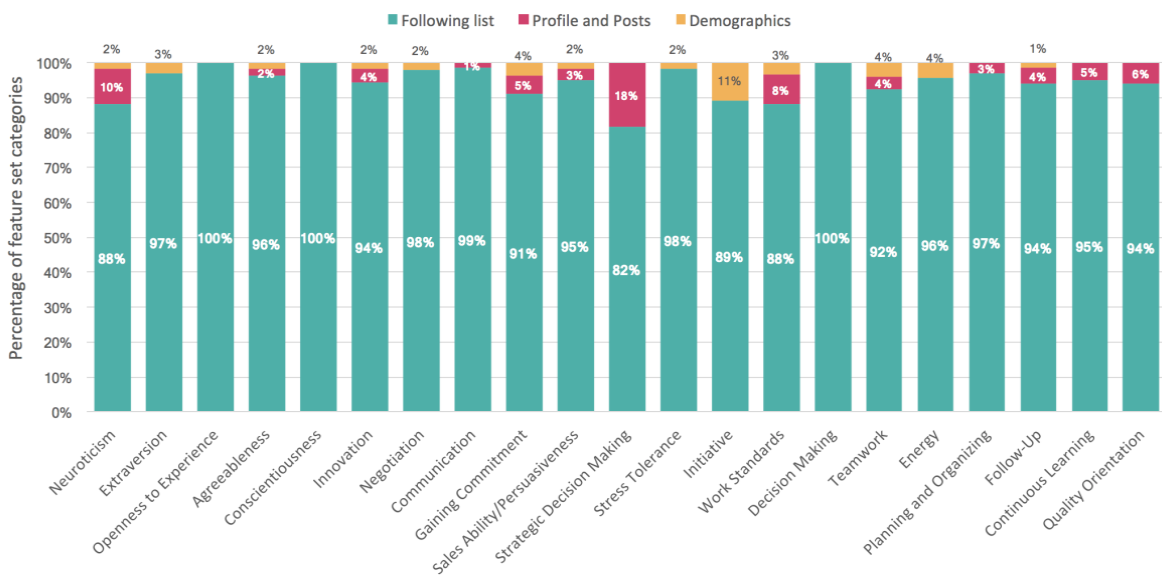

**Fig 2** Percentage of the model input feature set categories after feature selection in the three-level classification

The "Following List" appears to be the most influential feature set category as it covers 95% of the model input features after feature selection on average. This means that most of the significantly correlated features were among the Instagram following list as it was described in sect. 4.1. The number of input features was 56 on average after feature selection.

The performance results of all the experiments are summarized in **Table 5** and *Table 6*. Results of two-level classification indicate that based on accuracy, AUC, and precision the DL algorithm usually performs better, followed by GLM algorithm. DL outperforms other algorithms with respect to all three performance measurements (accuracy, AUC, and precision) in predicting Extraversion (74%, 0.802, 74%) Openness to Experience (69%, 0.681, 64%), Agreeableness (77%, 0.824, 70%), Gaining Commitment (70%, 0.744, 72%), Strategic Decision Making (73%, 0.798, 74%), Stress Tolerance (72%, 0.752, 73%), Initiative (76%, 0.832, 77%), Decision Making (74%, 0.798, 74%), Follow-up (70%, 0.784, 69%), and Quality Orientation (70%, 0.774, 67%). DL, also shows better results in terms of accuracy and AUC in predicting Communication (70%, 0.792), better accuracy and precision in predicting Sales Ability/Persuasiveness (65%, 63%) and Continuous Learning (74%, 70%), higher AUC for Conscientiousness (0.721), and better accuracy for Innovation (75%).

GLM delivers better results (accuracy, AUC, and precision) in predicting Negotiation (68%, 0.772, 71%), Energy (72%, 0.767, 73%), and Planning and Organizing (65%, 0.720, 62%). Based on accuracy and precision, GLM performs better in predicting Conscientiousness (66%, 65%). Based on accuracy and AUC, GLM has higher results for Continuous Learning (74%, 0.751). The Best AUC for Neuroticism (0.797) and Sales Ability/Persuasiveness (0.605), the best precision for Agreeableness (70%) and Communication (74%), and the most satisfactory accuracy for Initiative (76%), were among the highlights of the GLM algorithm.

The other two algorithms could show the best performance in predicting only one trait based on accuracy, AUC, and precision: LR in predicting Work Standards (71%, 0.731, 72%) and RF in predicting teamwork (72%, 0.743, 68%). Precisionwise, LR outperformed the other algorithms in predicting Innovation (74%), Stress Tolerance (73%), Follow-up (69%), and Neuroticism (79%). For the latest trait, it showed the best accuracy as well (76%).

Furthermore, results of three-level classification indicate that, the DL algorithm outperforms GLM and RF by showing higher accuracies and weighted $F_1$ scores in predicting most of the traits. DL performed better in predicting all five personality traits (accuracy and weighted $F_1$ score) namely Neuroticism (70%, 63%), Extraversion (70%, 66%), Openness to Experience (72%, 66%), Agreeableness (75%, 68%), and Conscientiousness (76%, 68%) and six out of sixteen soft skills namely Negotiation (71%, 68%), Gaining Commitment (71%, 65%), Strategic Decision Making (76%, 72%), Stress Tolerance (62%, 55%, similar with RF), Initiative (76%, 70%), and Planning and Organizing (76%, 71%).

Moreover, RF algorithm delivered the highest accuracy and weighted $F_1$ score in predicting six of the soft skills: Innovation (72%, 64%), Communication (72%, 65%), Stress Tolerance (62%, 55%), Work Standards (75%, 65%), Decision Making (61%, 50%), and Follow-up (71%, 62%).

Finally, GLM produced higher accuracy and weighted $F_1$ score in predicting four of the soft skills: Sales Ability/Persuasiveness (68%, 61%), Energy (71%, 66%), Continuous Learning (71%, 64%), Quality Orientation (67%, 60%). GLM (76%, 71%) and DL (75%, 72%) performed mutually good in predicting Teamwork.

Moreover, it can be seen from the results in **Table 5** that all four algorithms outperform the baseline model except for a few cases: LR in predicting Extraversion based on accuracy metric, LR in predicting Planning and Organizing (accuracy and precision), and LR in predicting Innovation (AUC). Also, the DL algorithm falls short in predicting Planning and Organization (accuracy and precision). However, these algorithms could produce better results with different feature sets. We used the same feature set for all the algorithms while predicting each trait and presented the results when we reached the best performance with at least one of the algorithms. So, the mentioned algorithms had slightly better performances that even outperformed the baseline model, yet to offer the results with identical sets of features, we overlooked those.

**Table 5**

*Accuracy, AUC, and Precision results for soft skill and personality trait prediction using GLM, LR, DL, and RF algorithms in two-level classification*

| Two-Level Classification | | | | | | |
|---|---|---|---|---|---|---|
| Traits | Performance | DT(Baseline) | LR | GLM | DL | RF |
| Neuroticism | Accuracy | 52% | **76%** | 72% | 67% | 67% |
|  | AUC | 0.479 | 0.773 | **0.797** | 0.771 | 0.765 |
|  | Precision | 53% | **79%** | 69% | 65% | 63% |
| Extraversion | Accuracy | 55% | 54% | 68% | **74%** | 56% |
|  | AUC | 0.529 | 0.629 | 0.746 | **0.802** | 0.731 |
|  | Precision | 54% | 57% | 68% | **74%** | 55% |
| Openness to Experience | Accuracy | 60% | 64% | 64% | **69%** | 65% |
|  | AUC | 0.594 | 0.630 | 0.651 | **0.681** | 0.633 |
|  | Precision | 58% | 61% | 60% | **64%** | 61% |
| Agreeableness | Accuracy | 52% | 62% | 73% | **77%** | 63% |
|  | AUC | 0.514 | 0.637 | 0.788 | **0.824** | 0.724 |
|  | Precision | 51% | 61% | **70%** | **70%** | 59% |
| Conscientiousness | Accuracy | 53% | 57% | **66%** | 65% | 65% |
|  | AUC | 0.498 | 0.591 | 0.702 | **0.721** | 0.698 |
|  | Precision | 56% | 59% | **65%** | 63% | 62% |
| Innovation | Accuracy | 63% | 65% | 71% | **75%** | 71% |
|  | AUC | 0.582 | 0.468 | 0.660 | 0.673 | **0.704** |
|  | Precision | 63% | **74%** | 67% | 70% | 68% |
| Negotiation | Accuracy | 53% | 66% | **68%** | 67% | 62% |
|  | AUC | 0.502 | 0.720 | **0.772** | 0.720 | 0.720 |
|  | Precision | 53% | 68% | **71%** | 65% | 59% |
| Communication | Accuracy | 51% | 59% | 65% | **70%** | 61% |
|  | AUC | 0.490 | 0.678 | 0.756 | **0.792** | 0.738 |
|  | Precision | 51% | 67% | **74%** | 66% | 57% |
| Gaining Commitment | Accuracy | 53% | 65% | 55% | **70%** | 62% |
|  | AUC | 0.488 | 0.703 | 0.720 | **0.744** | 0.667 |
|  | Precision | 53% | 67% | 56% | **72%** | 59% |
| Sales Ability/Persuasiveness | Accuracy | 55% | 60% | 60% | **65%** | 59% |
|  | AUC | 0.492 | 0.596 | **0.605** | 0.604 | 0.528 |
|  | Precision | 56% | 61% | 61% | **63%** | 59% |
| Strategic Decision Making | Accuracy | 58% | 70% | 67% | **73%** | 62% |
|  | AUC | 0.543 | 0.674 | 0.753 | **0.768** | 0.682 |
|  | Precision | 60% | 67% | 66% | **71%** | 62% |
| Stress Tolerance | Accuracy | 61% | 68% | 66% | **72%** | 68% |
|  | AUC | 0.539 | 0.700 | 0.730 | **0.752** | 0.677 |
|  | Precision | 61% | **73%** | 67% | **73%** | 66% |
| Initiative | Accuracy | 46% | 60% | **76%** | **76%** | 55% |
|  | AUC | 0.438 | 0.614 | 0.827 | **0.832** | 0.633 |
|  | Precision | 50% | 62% | 76% | **77%** | 56% |
| Work Standards | Accuracy | 56% | **71%** | 63% | 66% | 58% |
|  | AUC | 0.503 | **0.731** | 0.730 | 0.727 | 0.649 |
|  | Precision | 57% | **72%** | 63% | 63% | 59% |
| Decision Making | Accuracy | 49% | 62% | 62% | **74%** | 64% |
|  | AUC | 0.487 | 0.677 | 0.752 | **0.798** | 0.705 |
|  | Precision | 49% | 63% | 60% | **74%** | 59% |
| Teamwork | Accuracy | 55% | 64% | 71% | 66% | **72%** |
|  | AUC | 0.489 | 0.619 | 0.723 | 0.633 | **0.743** |
|  | Precision | 56% | 62% | 67% | 64% | **68%** |
| Energy | Accuracy | 56% | 59% | **72%** | 70% | 65% |
|  | AUC | 0.503 | 0.600 | **0.767** | 0.757 | 0.712 |
|  | Precision | 56% | 63% | **73%** | 68% | 62% |
| Planning and Organizing | Accuracy | 59% | 50% | **65%** | 56% | 64% |
|  | AUC | 0.519 | 0.559 | **0.720** | 0.668 | 0.716 |
|  | Precision | 58% | 55% | **62%** | 57% | 61% |
| Follow-Up | Accuracy | 53% | 62% | 66% | **70%** | 56% |
|  | AUC | 0.488 | 0.632 | 0.684 | **0.784** | 0.638 |
|  | Precision | 53% | **69%** | 67% | **69%** | 56% |
| Continuous Learning | Accuracy | 53% | 62% | **74%** | **74%** | 70% |
|  | AUC | 0.488 | 0.653 | **0.751** | 0.727 | 0.737 |
|  | Precision | 53% | 63% | 68% | **70%** | 65% |
| Quality Orientation | Accuracy | 51% | 65% | 62% | **70%** | 61% |
|  | AUC | 0.491 | 0.717 | 0.763 | **0.774** | 0.730 |
|  | Precision | 52% | 65% | 61% | **67%** | 58% |

In each row, the highest Accuracy, AUC, and Precision are typeset in bold

In addition, *Table 6* shows that we could produce models that outperform the baseline model for all the traits. Nevertheless, some algorithms which are fed with the same feature set do not always outperform the baseline. In some cases, they produce the same results as the baseline: RF in predicting Extraversion (accuracy), RF in predicting Agreeableness (accuracy and weighted F1 score), RF in predicting initiative (both metrics), RF in predicting Follow-up and Continues Learning (accuracy) and RF in predicting Quality Orientation (both metrics). DL also yields the same weighted F1 score as the baseline model in predicting work standards. In some other cases, the algorithms performed weaker than the baseline model: DL in predicting innovation and Follow-up (both metrics), DL in predicting communication, Sales ability, Work Standards, and Quality Orientation (accuracy), GLM in predicting Work Standards and Follow-Up (both metrics), RF in predicting Teamwork and Planning and Organizing (both metrics), and RF in predicting Gaining Commitment (accuracy). As mentioned above, these models could perform better with a slightly different feature set, yet we showed all the algorithms' results based on the same feature set that could produce the optimum results for at least one of the algorithms in predicting each trait.

Overall, the DL and GLM algorithms outperformed the others in most attempts to predict soft skills and personality. Two-level classifications displayed better results compared with three-level classifications. The class distribution is imbalanced in the three-level classification because of having more results in the medium range; thus accuracy metric is flawed to some extent. Considering weighted F1 score as the main metric for the three-level classification, the results are not as potent as those for the two-level classification.

Besides, some algorithms could not outperform the baseline model in the three-level problems. However, comparing these two classifications can improve the overall prediction accuracy. For instance, knowing that both classifications predicted a "high" Teamwork score for the same record increases confidence.

Agreeableness, Neuroticism, and Extraversion, among personality traits, and Initiative, Innovation, Decision Making, and Continuous Learning among soft skills are the easiest to predict in two classes of high or low from Instagram profiles. While, Agreeableness and Conscientiousness among personality traits and Strategic Decision Making, Initiative, Teamwork, and Planning and Organizing among behavioral competencies showed to be the most straightforward for predicting in three classes high, medium, and low.

*Table 6*
*Accuracy and Weighted F1 Score results for soft skill and personality trait prediction using GLM, DL, and RF algorithms in three-level classification*

| Three-Level Classification | | | | | |
|---|---|---|---|---|---|
| Traits | Performance | DT(Baseline) | GLM | DL | RF |
| Neuroticism | Accuracy | 63% | 69% | **70%** | 64% |
| | Weighted F1 Score | 48% | 62% | **63%** | 51% |
| Extraversion | Accuracy | 61% | 65% | **70%** | 61% |
| | Weighted F1 Score | 48% | 55% | **66%** | 50% |
| Openness to experience | Accuracy | 64% | 66% | **72%** | 66% |
| | Weighted F1 Score | 51% | 55% | **66%** | 54% |
| Agreeableness | Accuracy | 69% | 74% | **75%** | 69% |
| | Weighted F1 Score | 56% | 66% | **68%** | 56% |
| Conscientiousness | Accuracy | 70% | 75% | **76%** | 72% |
| | Weighted F1 Score | 59% | 66% | **68%** | 62% |
| Innovation | Accuracy | 65% | 68% | 57% | **72%** |
| | Weighted F1 Score | 52% | 63% | 50% | **64%** |
| Negotiation | Accuracy | 62% | 68% | **71%** | 68% |
| | Weighted F1 Score | 51% | 63% | **68%** | 61% |
| Communication | Accuracy | 66% | 69% | 64% | **72%** |
| | Weighted F1 Score | 61% | **65%** | 62% | **65%** |
| Gaining Commitment | Accuracy | 65% | 69% | **71%** | 64% |
| | Weighted F1 Score | 52% | 64% | **65%** | 53% |
| Sales Ability/Persuasiveness | Accuracy | 64% | **68%** | 60% | 65% |
| | Weighted F1 Score | 50% | **61%** | 54% | 53% |
| Strategic Decision Making | Accuracy | 69% | 72% | **76%** | 72% |
| | Weighted F1 Score | 58% | 69% | **72%** | 62% |
| Stress Tolerance | Accuracy | 55% | 60% | **62%** | **62%** |
| | Weighted F1 Score | 41% | **55%** | **55%** | **55%** |
| Initiative | Accuracy | 71% | 72% | **76%** | 71% |
| | Weighted F1 Score | 59% | 64% | **70%** | 59% |
| Work Standards | Accuracy | 73% | 68% | 68% | **75%** |
| | Weighted F1 Score | 64% | 61% | 64% | **65%** |
| Decision Making | Accuracy | 59% | 60% | **61%** | **61%** |
| | Weighted F1 Score | 43% | 49% | 49% | **50%** |
| Teamwork | Accuracy | 68% | **76%** | 75% | 66% |
| | Weighted F1 Score | 56% | 71% | **72%** | 54% |
| Energy | Accuracy | 62% | **71%** | 66% | 65% |
| | Weighted F1 Score | 51% | **66%** | 62% | 55% |
| Planning and Organizing | Accuracy | 69% | 70% | **76%** | 68% |
| | Weighted F1 Score | 56% | 64% | **71%** | 55% |
| Follow-Up | Accuracy | **71%** | 61% | 66% | **71%** |
| | Weighted F1 Score | 60% | 55% | 59% | **62%** |
| Continuous Learning | Accuracy | 68% | **71%** | 70% | 68% |
| | Weighted F1 Score | 56% | **64%** | 63% | 56% |
| Quality Orientation | Accuracy | 66% | **67%** | 59% | 66% |
| | Weighted F1 Score | 52% | **60%** | 55% | 52% |

In each row, the highest Accuracy and Weighted F1 Score are typeset in bold

# 5. Discussion, conclusion, and future directions

This study showed that job-related information like soft skills and personality traits are accurately and unobtrusively predictable through computational models based on Instagram profiles. Initiative, Innovation, Decision Making, and Continuous Learning among soft skills and Agreeableness, Neuroticism, and Extraversion, among personality traits, are the easiest to predict from Instagram profiles in two-level classifications. Whereas Strategic Decision Making, Initiative, Teamwork, and Planning and Organizing among behavioral competencies and Agreeableness and Conscientiousness among personality traits were shown to be the most straightforward for predicting in three-level classifications.

Using an Iranian sample, we showed that despite different cultural implications from western countries, Iranian users with similar soft skills and personality traits also display similar behavioral patterns on social media. These patterns can be explored to find job candidates' scores in the job market in-demand soft skills and personality traits and to make an immense talent pool.

We performed several machine learning algorithms (generalized linear model, logistic regression, deep learning, and random forest) on data in two-class and three-class problems. Based on the experiments, deep learning models mostly outperformed in both two-class and three-class problems by demonstrating 70% and 69% average Accuracy respectively, followed by generalized linear models. Predicting numerical scores was beyond the purpose of this study which can be investigated in future attempts.

We used correlation analysis (Pearson correlation) for the feature selection procedure. Applying other feature selection methods to investigate the most relevant features further and improve the performance is a future track to dig deeper.

Other than soft skills or behavioral competencies and personality traits which are extremely influential in career success and therefore closely targeted by recruiters in job candidates, there are other influential behavioral traits like values and attitudes that can be the target of future studies.

In this study, we showed that the most predictive features were the Instagram users' following lists. Although we did not separately analyze the predictive power of each feature set category (following lists, profile and post parameters, and demographic features), the feature selection results based on correlation analysis showed that most of the significantly correlated features were among the Instagram following lists. The conceptual content of photos, videos, texts, and likes of Instagram users has not been investigated in this study, which can be analyzed in other AI-based recruitment experiments.

Facebook Likes were previously shown to have predictive power regarding personality and other personal attributes (Kosinski et al., 2013). The similarity between Facebook Likes and Instagram Following list and the results of the present study further validates the power of digital records in predicting personal traits. All the other online digital records of the job candidates, which are now as important as their offline reputation, can be the subject of future investigations. Digital footprints can be active or passive. Logging in to a website or subscribing to a newsletter are examples of active digital footprints, while being documented by websites on the number of our visits to their page or being scanned on our browsing history by some websites are forms of passive digital footprints. Examples of online environments where digital footprints are generated and can be investigated for recruitment and selection purposes in addition to social media are online shops, online financial services, online reading and News websites, health and fitness apps and websites, and online gaming and entertainment platforms. Screening job candidates should be implemented with respect to ethical and legal concerns on the one hand and their intentions to apply and join the companies using such data and methods on the other hand. Several previous studies have already investigated these aspects; however, validation of such practices has not been examined in most of the online environments mentioned above and, therefore, is an open path for future works.

By using digital footprints as a screening tool, we can target not only active job seekers but qualified passive job seekers to create large talent pools. Screening can start even younger, and children with extraordinary talents can be targeted. Companies using this technique can become leaders in their industry using proper attraction strategies and allocating learning and development costs for these emerging talents to become experts in the most fitting occupations. The spending will be compensated by the reduction in the cost of developing expert employees who lack soft skills or any in-demand skill apart from the increased revenue as a result of attracting the best talent. Educational and career counseling Institutions can also analyze their customers' online activity to understand them more profoundly and provide more effective counseling. The validation of the noted benefits can be the subject of further examinations.

**Funding**

No funding.

**Data availability**

Data will be made available on request.